\documentclass[preprint, 12pt]{amsart}
\usepackage[table]{xcolor}
\usepackage{times}
\usepackage{amssymb, amsmath}
\usepackage{amsthm}
\usepackage{tikz}
\usepackage{bm}
\usetikzlibrary{matrix,arrows}
\usepackage{enumitem}
\usepackage{booktabs}

\theoremstyle{plain}

\theoremstyle{remark}

\definecolor{Gray}{gray}{0.95}
\newcolumntype{g}{>{\columncolor{Gray}}c}

\graphicspath{{./images/}}

\begin{document}
\title{Gamma distribution-based sampling for imbalanced data}
\author{Firuz Kamalov$^1$$^{*}$ and  Dmitry  Denisov$^2$}

\address{$^{1}$ Canadian University Dubai, Dubai, UAE.}
\email{\textcolor[rgb]{0.00,0.00,0.84}{firuz@cud.ac.ae}}

\address{$^{2}$ Deloitte, Dubai, UAE}
\email{\textcolor[rgb]{0.00,0.00,0.84}{dmdenisov@deloitte.com}}

\date{May 15, 202
\newline \indent $^{*}$ Corresponding author}

\begin{abstract}
Imbalanced class distribution is a common problem in a number of fields including medical diagnostics, fraud detection, and others. It causes bias in classification algorithms leading to poor performance on the minority class data. In this paper, we propose a novel method for balancing the class distribution in data through intelligent resampling of the minority class instances. The proposed method is based on generating new minority instances in the neighborhood of the existing minority points via a gamma distribution. Our method offers a natural and coherent approach to balancing the data. We conduct a comprehensive numerical analysis of the new sampling technique. The experimental results show that the proposed method outperforms the existing state-of-the-art methods for imbalanced data. Concretely, the new sampling technique produces the best results on 12 out of 24 real life as well as synthetic datasets. For comparison, the SMOTE method achieves the top score on only 1 dataset. We conclude that the new technique offers a simple yet effective sampling approach to balance data.
\end{abstract}

\maketitle


\section{Introduction}
Imbalanced class distribution refers to a setting where  one class label significantly outnumbers another class label in the dataset. It occurs in a number of applications including medical diagnostics, insurance fraud, spam email, and others that involve rare events \cite{Krawczyk, Mani, Triguero}. Imbalanced class distribution leads to bias in classification algorithms. Since the goal of a classification algorithm is to maximize the overall predictive accuracy it would direct most of its efforts to correctly classifying the majority label data. If the payoff from correctly classifying the minority instances is not significant enough the classifier will likely ignore such points. At the same time, correctly classifying the minority points is often of more importance than the majority points. For instance, identifying a patient with a rare disease is far more crucial than identifying a healthy individual.

There are a number of approaches to dealing with imbalanced data. The existing approaches can be broadly categorized as cost sensitive classification, one-class classification, resampling, and blended methods. In cost sensitive classification, the minority class instances are assigned a higher weight than the majority class instances forcing the classifier to focus more on the minority data. In one-class classification, the minority points are considered separately from the majority points. In resampling methods, the data is balanced through undersampling the majority class data or oversampling the minority class data. Finally, blended methods combine two or more approaches to address class imbalance.
Although the existing solutions perform well in certain cases, they suffer from several drawbacks. Cost sensitive approaches are vulnerable to overfitting as narrow regions in space are endowed with higher weights. One-class methods do not take the full advantage of the available data. Since each target class is considered independently any possible interactions between classes are ignored. Similarly, undersampling methods do not utilize all the available information as only a fraction of the majority class is used in classification. Oversampling techniques seem to avoid the above issues by carefully creating new minority class points to balance the data. However, the assumptions underlying oversampling techniques may not necessarily be valid for a given dataset.
There remains a considerable room for improvement in the existing oversampling methods. Blended approaches combine the above methods and  therefore inherit their drawbacks.

In this paper, we propose a new technique to balance skewed class distribution through intelligent oversampling of the minority class points. The proposed method uses the gamma distribution to create new minority class points. Concretely, given an existing minority class point $p_0$ we generate a new minority point  in the neighborhood of $p_0$ using the gamma distribution. The use of the gamma distribution allows for the new minority points to be generated close to the existing minority points. We believe that such approach is more natural than other similar oversampling techniques that use the uniform distribution. In addition, since the amma distribution is asymmetric, we can target the locations of the new minority points in the most likely directions based on the neighboring minority points.

We conduct a series of experiments to test the efficacy of the proposed oversampling technique. The performance of the proposed method is benchmarked against a number of popular oversampling techniques such as ROS, SMOTE and ADASYN.
First, we use simulated data to illustrate that the proposed method can significantly outperform the existing sampling techniques. Second, we analyze the performance of the new method on a range of real life datasets (Table \ref{info}) with different imbalance ratios. The results of the experiments show that the proposed sampling method consistently outperforms the competitor approaches. Given the variety of datasets that were used in our experiments, we conclude that the proposed method would be effective in a range of applications.

Our paper is structured as follows. In Section 2, we survey the existing approaches to deal with imbalanced data. In Section 3, we describe the details of the proposed method. In Section 4, we present the results of the numerical experiments performed to test the efficacy of the proposed method. We conclude with a brief summary of the paper in Section 5. 

\section{Literature}

The problem of imbalanced data has been well covered in the literature \cite{Thabtah}. There are exist several approaches to dealing with skewed class distribution including cost sensitive classification \cite{Cheng, Khan} , one-class classification \cite{Xu}, resampling\cite{Fernandez}, and blended methods \cite{Cao, Haix}. Among the existing variety of approaches to combat class imbalance sampling techniques have gained the biggest acceptance.

Sampling techniques - and in particular oversampling techniques - have received greater attention in the literature than other methods due to their recent success. One of the most popular oversampling approaches called SMOTE was proposed by Chawla et al. in \cite{Chawla}. SMOTE operates by linearly interpolating new minority points between existing neighboring minority points. In other words, the new points are randomly interspersed along the lines connecting neighboring minority class points. An extension of SMOTE called ADASYN was proposed by He et al. in \cite{He, He2}. ADASYN generates the new minority points in the same fashion as SMOTE except with a greater emphasis on the border regions between the majority and minority classes. The regions that are close to the border with the majority class data are assigned a greater number of new minority points. A detailed study of SMOTE and its properties can be found in \cite{Elreedy}. 
Modern oversampling approaches use more sophisticated techniques to generate new points. The authors in \cite{Kamalov1} apply kernel density estimation to approximate the distribution of the minority points. The resulting estimator is used to generate new minority points. In \cite{Mathew}, the authors propose to overcome the linear nature of SMOTE algorithm for nonlinear problems by sampling from the feature space of kernel SVM classifier. 

Undersampling techniques work by sampling a subset of the majority class to balance the class distribution \cite{Lin, Liu}. Undersampling techniques are particularly popular in bagging methods. In bagging, subsets of the majority class are repeatedly sampled and base classifiers are trained on the balanced datasets. Then an ensemble rule is used to determine the output of the combined classifier. Ensemble techniques that use bootstrap undersampling are a popular approach to imbalanced data \cite{Galar, Wang}.  

The gamma distribution is a popular statistical tool used in a number of applications \cite{Krishnamoorthy}. In \cite{Zhang}, the authors use multivariate gamma distribution to investigate the performance of radio frequency and optical wireless communication systems. Gamma distribution and its extensions have also been  used to model a variety of data and processes \cite{Belikov, Bourguignon, Klakattawi}.
It is also used as a kernel function in  nonparametric density estimation. Standard symmetric kernels for density estimation do not perform well with respect to positive or bounded variables. The issue is particularly acute in positive variables with a large probability mass near zero. Gamma kernels can be used effectively to estimate such distributions \cite{Chen, Malec}.   
In \cite{Bouezmarni}, authors use the positive asymmetry of the gamma distribution to model highly skewed income distribution. The authors in \cite{Jeon} apply a gamma kernel to estimate insurance loss distribution.


\section{Methodology}

\begin{figure}[h!]
\centering
\includegraphics[scale=0.5]{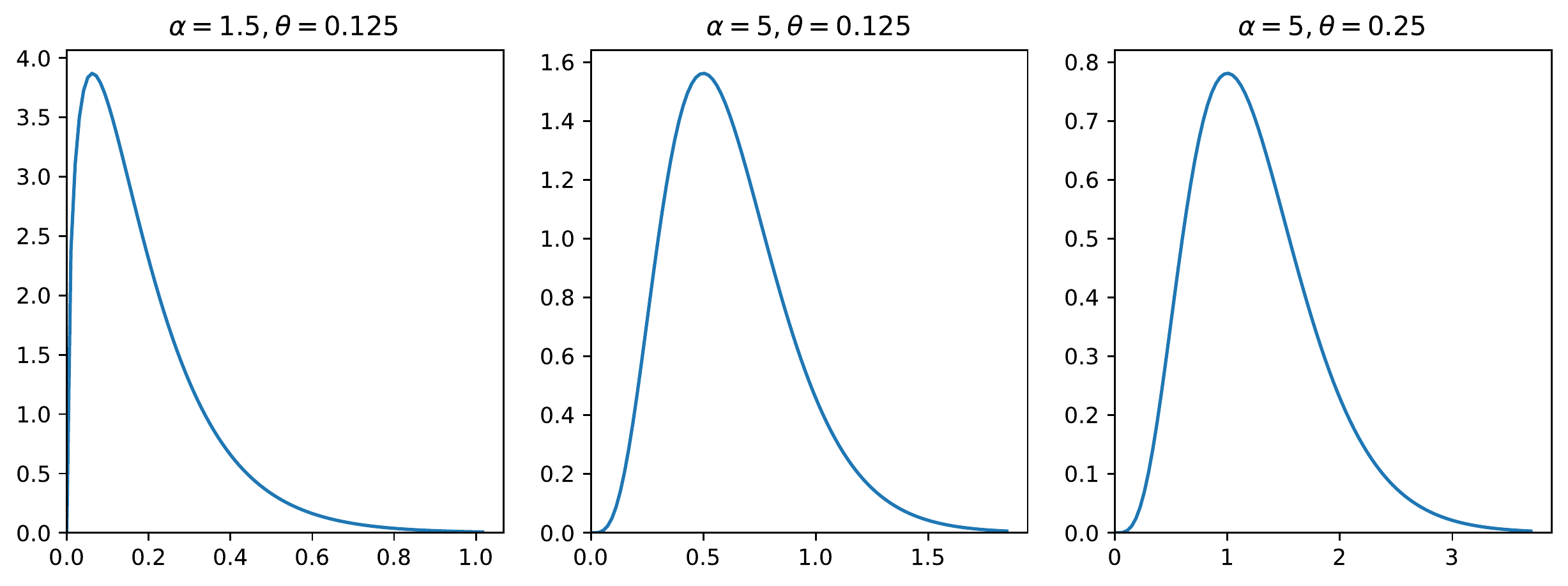}
\caption{The effects of varying $\alpha$ and $\theta$ values on the shape of the distribution.}
\label{gamma_example}
\end{figure}

In this section, we provide the necessary background and describe the algorithm for the proposed method. The proposed sampling technique is based on  univariate gamma distribution. The gamma distribution is  a two-parameter family of continuous distributions. It is often used in Bayesian statistics as prior conjugate distribution for various types of rate parameters. It arises in processes involving the waiting times between Poisson distributed events. The probability density function for a gamma distribution is given by the equation
\begin{equation}
f(x; \alpha, \theta) = \frac{x^{\alpha-1}e^{-\frac{x}{\theta}}}{\Gamma(\alpha)\theta^{\alpha}},
\label{gamma_eq}
\end{equation}
where $\Gamma(z) = \int_0^\infty x^{z-1}e^{-x} \, dx$ is the gamma function, $\alpha$ is the shape parameter, and $\theta$ is the scale parameter. The parameter $\alpha$ controls the shape of the distribution. As shown in Figure \ref{gamma_example}, an increase in the value of $\alpha$ leads to a more symmetrical gamma distribution. The parameter $\theta$ controls the scale of the distribution. An increase in the value of $\theta$ results in an increase in the scale of the  distribution. Note there exists an alternative notation in the literature for the scale parameter $\theta$, whereby it is replaced by the rate parameter $\lambda =\frac{1}{\theta}$. The resulting equation is given by 
$f(x; \alpha, \lambda) = \frac{\lambda^{\alpha}x^{\alpha-1}e^{-\lambda x}}{\Gamma(\alpha)}$. It is also worth noting that the exponential distribution is a special case of the gamma distribution with $\alpha=1$. Similarly, the $\chi^2$ is another special case of the gamma distribution.

It is an easy exercise in calculus to show that the maximum value of $f(x; \alpha, \theta)$ occurs at 
\begin{equation}
\label{max}
x = \theta(\alpha-1).
\end{equation}
For instance, given a gamma distribution with parameters $\alpha = 5$ and $\theta = \frac{1}{8}$ the maximum value of the pdf occurs at $x=0.5$ (Figure \ref{gamma_example}).

In our proposed sampling procedure we are guided by the following  two core principals:
\begin{enumerate}
\item The new minority class points must lie in the close vicinity of the existing minority points.
\item The new minority class points must be positioned in the direction of the neighboring minority points.
\end{enumerate}

\begin{figure}[h!]
\centering
\includegraphics[scale=0.5]{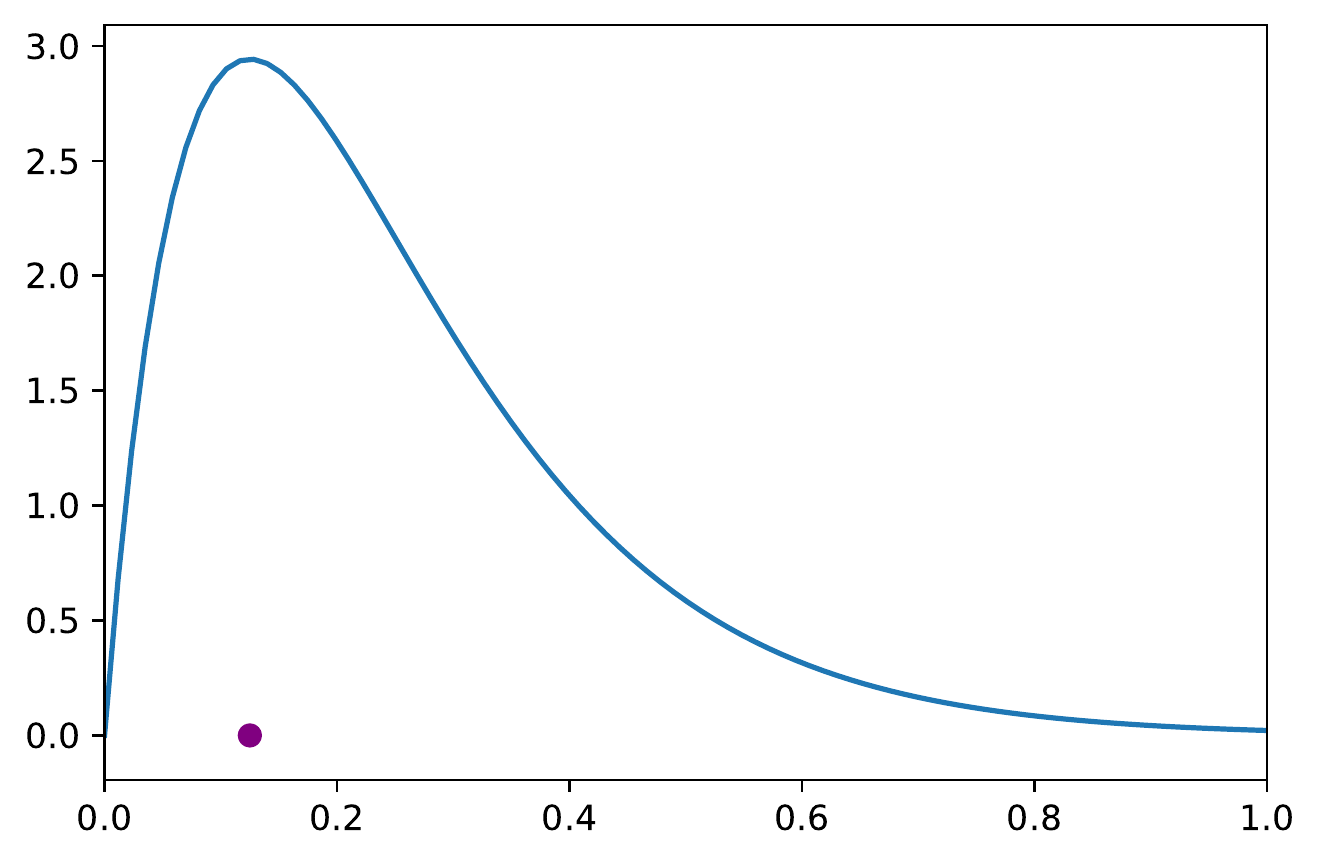}
\caption{The base gamma distribution is given by the parameters $\alpha=2, \theta =\frac{1}{8}$. The maximum value of the distribution occurs at $x=0.125$ (Equation \ref{max}). }
\label{base}
\end{figure}

To achieve the first goal above we place a base gamma distribution to each existing minority point so that the maximum value of the pdf occurs at the given minority point  (Figure \ref{base}). Accordingly, the new minority points will be more likely generated close to the chosen minority point.

\begin{figure}[h!]
\centering
\includegraphics[scale=0.5]{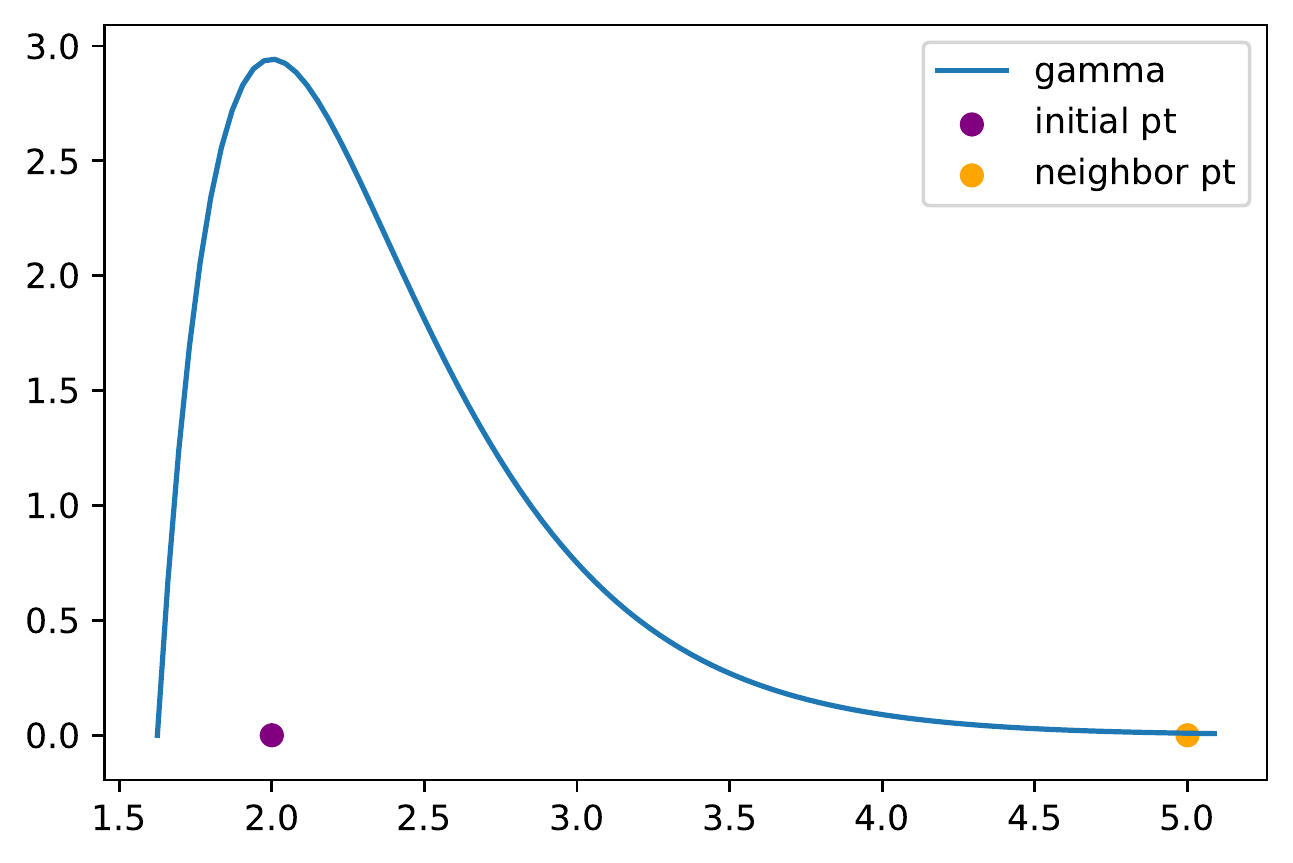}
\caption{Given a minority point at $x=2$ and a neighboring point at $x=5$ the base gamma distribution is scaled and oriented accordingly.}
\label{2d_demo}
\end{figure}

To achieve the second goal above we orient the base gamma distribution in the direction of neighboring minority points. The base gamma distribution is also scaled according to the distance to the neighboring point. In Figure \ref{2d_demo}, the initial  minority point is located at $x=2$ and the neighboring minority point is located at $x=5$. The new minority points will be generated according to the scaled gamma distribution. As shown in Figure \ref{2d_demo}, the new minority points are more likely to be generated near the initial minority point and in the direction of the neighboring minority point.

\begin{figure}[h!]
\centering
\includegraphics[scale=0.5]{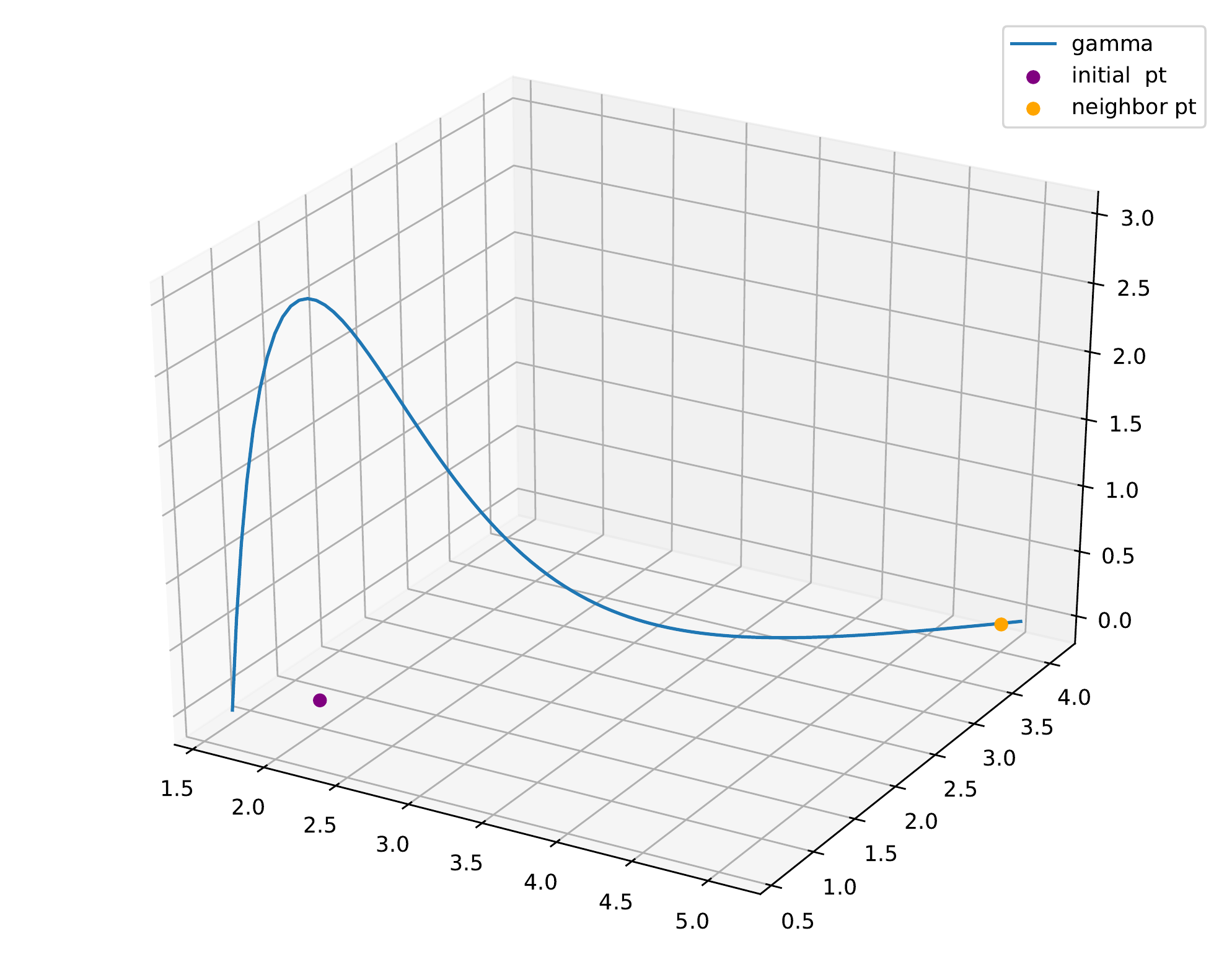}
\caption{Given a minority point at $(2, 1)$ and a neighboring point at $(5, 4)$ the base gamma distribution is scaled and oriented accordingly.}
\label{3d_demo}
\end{figure}

In Figure \ref{3d_demo}, we illustrate the gamma distribution for points in 2D. Given the initial minority point at $(2, 1)$ and a neighboring point at $(5, 4)$, we scale and orient the base gamma distribution  accordingly. Thus, the new minority points generated by the gamma distribution will be more likely to generated near the initial point $(2, 1)$ and in the direction of the neighboring point at $(5, 4)$. 

In summary, the proposed sampling technique starts by selecting an initial minority point $p$ and a neighboring minority point $p'$. Then, the base gamma distribution is scaled and oriented according to $p$ and $p'$ as illustrated in Figure \ref{3d_demo}. The new minority points will be generated according to the resulting gamma distribution. The proposed technique ensures that the new minority points a generated near the existing minority points and in the direction where they are most likely to appear based on the neighboring minority points. The details of the proposed algorithm are outlined  below.
\\
\\
\textbf{Algorithm}
\\
\line(1,0){150}
\\
\begin{enumerate}
\item Choose the parameters $\alpha$ and $\theta$ for the base gamma distribution (Equation \ref{gamma_eq}). 
\item Calculate the coordinates of the maximum value of the distribution using Equation \ref{max}. 
\item Randomly choose $N$ minority points (with replacement), where $N$ equals the difference between the number of majority and minority points. 

\item For each randomly chosen minority point $p$:
\begin{enumerate}[label=\roman*., itemsep=1ex]

\item Randomly choose one of the $k$ nearest neighbors $p'$. Define vector $v = p'-p$.

\item Generate a value $t$ using the base Gamma distribution. 

\item Define a new minority point as $q = p + (t-m)\cdot v$, where $t-m$ represents the scaling factor. 
\end{enumerate}

\item Repeat until the number of the minority points equals the number of majority points.
\end{enumerate}

\begin{figure}[h!]
\centering
\includegraphics[scale=0.7]{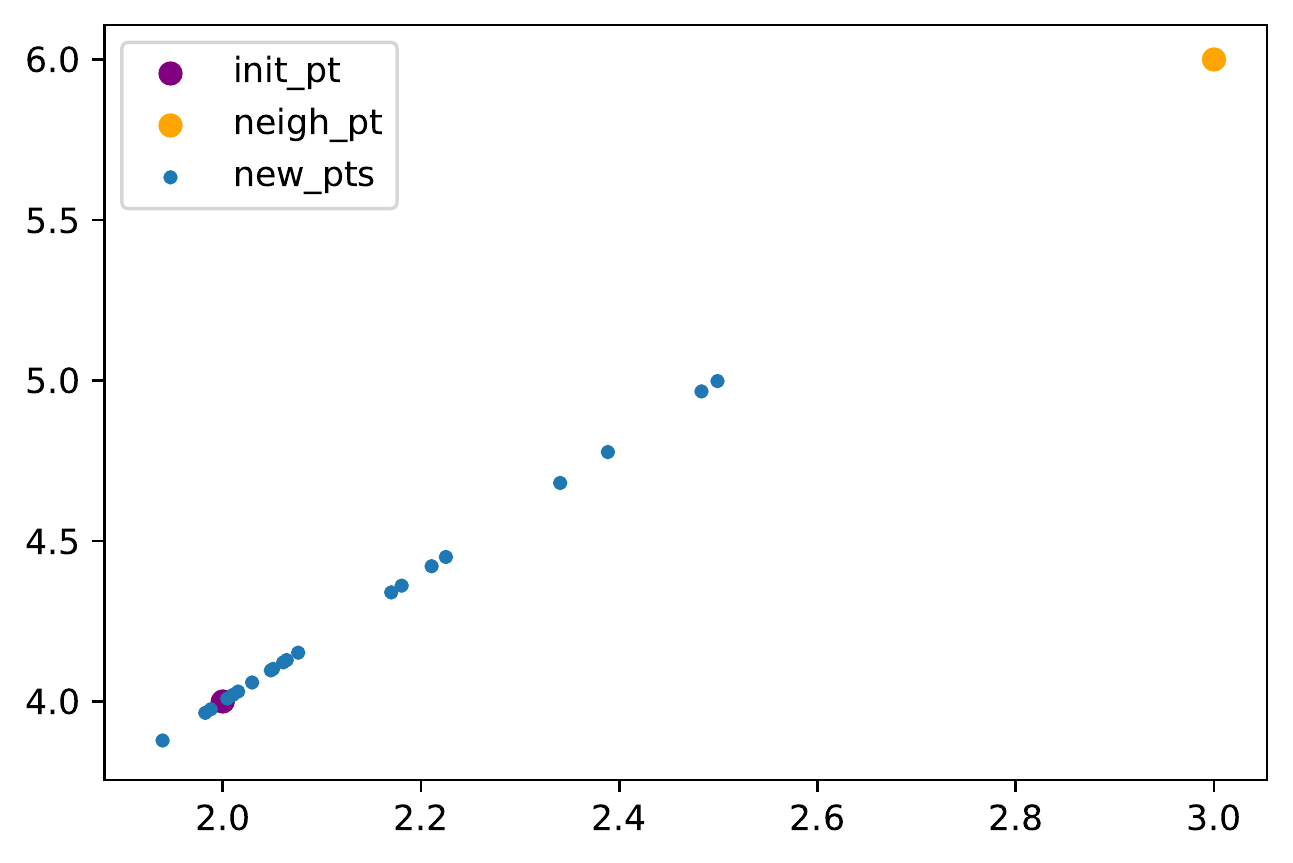}
\caption{Given a minority point at $x=2$ and a neighboring point at $x=5$ the base gamma distribution is scaled and oriented accordingly.}
\label{pts_gen}
\end{figure}

The process of generating new points using the the gamma distribution is illustrated in Figure \ref{pts_gen}. As shown in the figure, most of the new points are located near the initial minority point. In addition, a portion of the new points lies on the opposite side of the neighboring point. We find it natural to allow some of the points to be generated outside the range between the initial and neighboring points. By adjusting the shape and scale parameters we can control the process of generation of new points. The increase in the shape parameter would result in a more symmetrical distribution of the new points around the initial minority point. The increase in the scale parameter would result in points being more likely generated further from the initial point.


\section{Numerical Experiments}
In this section, we present the results of the numerical experiments that were carried out to test the performance of the proposed method. We benchmark the performance of the new method against several  popular sampling techniques including Random Undersampling (RUS), Random Oversampling (ROS), SMOTE, and ADASYN. The experiments are carried out using both simulated and real life data. The synthetic dataset is created to illustrate a scenario where the proposed sampling method would be advantageous. We also demonstrate the performance of the new method on a wide array of real life datasets. We employ a total of 24 real life datasets whose class ratio ranges from 8.6:1 to 130:1. The results show an improved performance of the new method over other existing sampling techniques.

We use 5-fold cross validation throughout our experiments. At each iteration, 80\% of the original (imbalanced) data is sampled and a classifier is trained on the resulting balanced dataset. Then the trained classifier is tested on the holdout (imbalanced) 20\% fold. 
The reported results represent the mean outcome over the 5 holdout sets. The number of nearest neighbors used in the sampling algorithms is $k=3$. The numerical experiments are implemented in Python using the \textit{sklearn} \cite{Pedregosa} and \textit{imblearn} \cite{Lemaitre} machine learning libraries.

Measuring the performance of a classifier in the case of imbalanced dataset requires a careful consideration. Traditional metrics such as accuracy and error rate do not fully reflect the performance of a classifier on the minority set. To this end, we use the $F_1$-score and AUC to measure the performance of sampling techniques. The $F_1$-score is the harmonic mean of precision and recall
$$F_1 = 2\cdot \frac{\mbox{precision}\cdot \mbox{recall}}{\mbox{precision}+ \mbox{recall}}.$$
The $F_1$-score indicates the ability of the classifier to efficiently identify the positive instances in the dataset. AUC is another commonly used metric for imbalanced data. It represents the probability that a classifier will rank a randomly chosen positive instance higher than a randomly chosen negative one.

\subsection{Simulated data}
\begin{figure}[h!]
\centering
\includegraphics[scale=0.7]{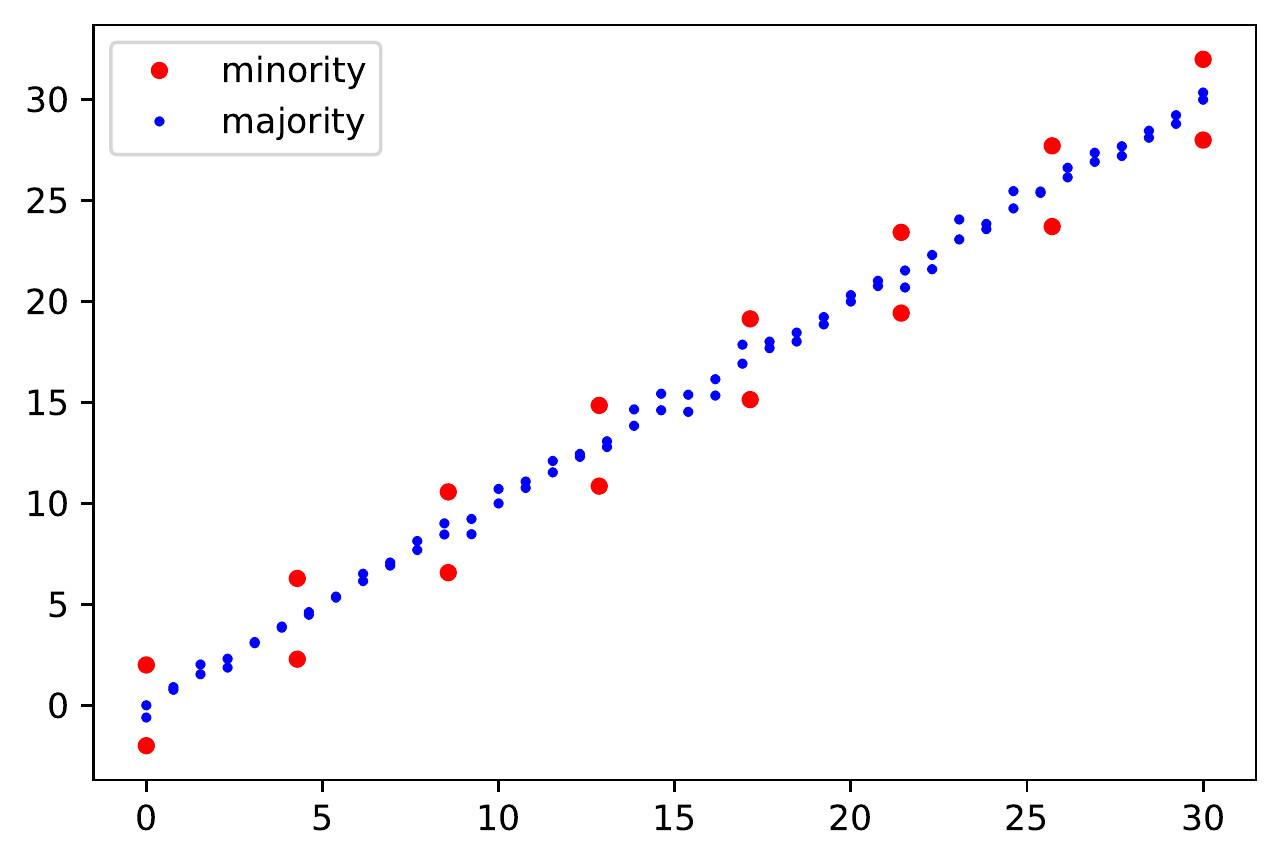}
\caption{Distribution of synthetically generated data.}
\label{syn_example}
\end{figure}

We use simulated data to demonstrate the efficacy of the proposed method. Concretely, we construct a dataset where the majority class points are randomly distributed along a straight line and the minority points are placed on  two sides of the majority set. A sample distribution of the data is illustrated in Figure \ref{syn_example}. In our experiment, we employ a synthetic imbalanced dataset consisting of 5,500 points of which 10\% belong to the minority class.  The goal of the experiment is to analyze the performance of gamma sampling compared to SMOTE sampling method. As a benchmark we also test the original imbalanced dataset.

We use  Nearest Neighbor (NN), Random Forest (RF), and Support Vector Machines (SVM) classifiers to compare the performance of the sampling methods. Each classifier is trained and tested on imbalanced, SMOTE-balanced, and Gamma-balanced datasets. The main measures of classifier performance are the $F_1$-score and AUC. The result of the experiment are presented in Table \ref{syn_table}. As can seen from the table the results are overwhelmingly in favor of the Gamma-based sampling. The Gamma-based method outperforms SMOTE and the original imbalanced datasets on every classifier both in terms of $F_1$-score and AUC values. Gamma-based sampling in conjunction with the SVM classifier produces the best overall results of $F_1 = 0.52$  and AUC = $0.33$.

\begin{table}[h!]
\centering
\caption{Experiments on synthetic data}
\label{syn_table}
\begin{tabular}{lllrrrr}
\toprule
Sampling & Classifier &  $F_1$-score &  precision &  recall &  AUC \\
\midrule
\rowcolor{Gray}
initial &   NN &    0.0042 &     0.2000 &  0.0021 &  0.0928 \\
smote &   NN &    0.1931 &     0.1229 &  0.7103 &  0.1069 \\
\rowcolor{Gray}
gamma &   NN &    \textbf{0.3105} &     0.5817 &  0.5514 &  \textbf{0.2334} \\
\hline
\rowcolor{Gray}
initial &   RF &    0.0044 &     0.2000 &  0.0022 &  0.0929 \\
smote &   RF &    0.0440 &     0.0307 &  0.0786 &  0.0863 \\
\rowcolor{Gray}
gamma &   RF &    \textbf{0.0535} &     0.0516 &  0.0558 &  \textbf{0.0888} \\
\hline
\rowcolor{Gray}
initial &  SVM &    0.0000 &     0.0000 &  0.0000 &  0.0909 \\
smote &  SVM &    0.1861 &     0.1609 &  0.2221 &  0.1076 \\
\rowcolor{Gray}
gamma &  SVM &    \textbf{0.5195} &     0.6071 &  0.4560 &  \textbf{0.3286} \\
\bottomrule
\end{tabular}
\end{table}

The results of the experiment on the synthetic dataset illustrate the weakness of SMOTE and other related sampling methods. Concretely, in case where  neighboring minority points are separated by a group of majority points the SMOTE algorithm is likely to generate new minority points in the midst of the majority points. Consequently, a classification algorithm would have difficulty distinguishing between the two classes. Our sampling method is designed to address this issue by employing the gamma distribution in place of uniform distribution. In the scenario presented in Figure \ref{syn_example}, using a gamma distribution would generate new minority points that are closer to the original minority points and therefore less likely to mix with the majority points. 

\subsection{Real life data}
We test the performance of the proposed method on a range of real life datasets. The datasets are chosen to represent varying domains and imbalance ratios (Table \ref{info}). The class ratios of the datasets used in the experiment range from 8.6:1 to 130:1. The number of samples and features in the datasets range from 336 to 20,000 and from 6 to 294 respectively. Thus, we obtain a comprehensive experimental analysis of the proposed method. 

\begin{table}[h!]
\centering
\caption{Details of the Experimental Datasets.}
\label{info}
\begin{tabular}{lrlllrr}
\toprule
ID &           Name &          Repository \& Target &                Ratio &     \#S &   \#F \\
\midrule
\rowcolor{Gray}
0  &      ecoli &             UCI, target: imU &                8.6:1 &    336 &    7 \\
1  &     optical\_digits &               UCI, target: $8$ &                9.1:1 &   5620 &   64 \\
\rowcolor{Gray}
2  &     satimage &               UCI, target: $4$ &                9.3:1 &   6435 &   36 \\
3  &      pen\_digits &               UCI, target: $5$ &                9.4:1 &  10992 &   16 \\
\rowcolor{Gray}
4  &     abalone &               UCI, target: $7$ &                9.7:1 &   4177 &   10 \\
5  &     sick\_euthyroid &  UCI, target: sick euthyroid &                9.8:1 &   3163 &   42 \\
\rowcolor{Gray}
6  &    spectrometer &            UCI, target: $>=44$ &             11:1 &    531 &   93 \\
7  &    car\_eval\_34 &    UCI, target: good, v good &             12:1 &   1728 &   21 \\
\rowcolor{Gray}
8  &    us\_crime &           UCI, target: $>0.65$ &             12:1 &   1994 &  100 \\
9  &        yeast\_ml8 &            LIBSVM, target: $8$ &             13:1 &   2417 &  103 \\
\rowcolor{Gray}
10 &      scene &   LIBSVM, target: $>$one label &             13:1 &   2407 &  294 \\
11 &       libras\_move &               UCI, target: $1$ &             14:1 &    360 &   90 \\
\rowcolor{Gray}
12 &      thyroid\_sick &            UCI, target: sick &             15:1 &   3772 &   52 \\
13 &         coil\_2000 &  KDD, CoIL, target: minority &             16:1 &   9822 &   85 \\
\rowcolor{Gray}
14 &       arrhythmia &              UCI, target: 06 &             17:1 &    452 &  278 \\
15 &    solar\_flare\_m0 &            UCI, target: $>0$ &             19:1 &   1389 &   32 \\
\rowcolor{Gray}
16 &             oil &        UCI, target: minority &             22:1 &    937 &   49 \\
17 &        car\_eval\_4 &           UCI, target: vgood &  26:1 &   1728 &   21 \\
\rowcolor{Gray}
18 &      wine\_quality &       UCI, wine, target: $<=4$ &  26:1 &   4898 &   11 \\
19 &        letter\_img &               UCI, target: Z &  26:1 &  20000 &   16 \\
\rowcolor{Gray}
20 &        yeast\_me2 &             UCI, target: ME2 &  28:1 &   1484 &    8 \\
21 &      ozone\_level &             UCI, ozone, data &  34:1 &   2536 &   72 \\
\rowcolor{Gray}
22 &      mammography &        UCI, target: minority &  42:1 &  11183 &    6 \\
23 &     abalone\_19 &              UCI, target: 19 &  130:1 &   4177 &   10 \\
\bottomrule
\end{tabular}
\end{table}

We benchmark the performance of the proposed method against a number of other traditional sampling techniques including SMOTE, ADASYN, Random Oversampling (ROS), and Random Undersampling (RUS). After sampling and balancing the datasets the Random Forrest classifier is applied to train and test on the data. We measure the performance of the classifier using the $F_1$-score and AUC. The $F_1$-score values are reported in Table \ref{f1}. As shown in the table, the proposed method (Gamma) produces the best result in 12 out of 24 datasets. For comparison, the SMOTE algorithm produces the best result just on 1 dataset. On certain datasets, the margin of performance is substantial - as in the case of the \textit{arrhythmia} dataset where the $F_1$-score of the proposed method is $0.17$ higher than the second highest score. The results show that the proposed method can be applied effectively to datasets with different class ratios. In particular, the \textit{ecoli} dataset has a class ratio of 8.6:1 and \textit{abalone\_19} has class ratio of 130:1. In addition, the new method produces strong results across different fields of application including medicine, image recognition, crime, and others.
Overall, the results in Table \ref{f1} show that the proposed method is superior to the benchmark sampling techniques. 

\begin{table}[h!]
\centering
\caption{Classification performance - as measured by the $F_1$-score - using various sampling techniques.}
\label{f1}
\begin{tabular}{llrrrrrr}
\toprule
{ID} &            Name &  Initial &   Gamma &   SMOTE &  ADASYN &  ROS &  RUS \\
\midrule
\rowcolor{Gray}
0  &           ecoli &   0.6353 &  \textbf{0.6555} &  0.6278 &  0.6101 &    0.4490 &    0.5896 \\
1  &  optical\_digits &   0.8767 &  \textbf{0.9386} &  0.9227 &  0.9121 &    0.9175 &    0.8966 \\
\rowcolor{Gray}
2  &        satimage &   0.6181 &  0.6537 &  0.6753 &  0.6828 &    0.6876 &    0.5653 \\
3  &      pen\_digits &   0.9789 &  \textbf{0.9863} &  0.9821 &  0.9789 &    0.9820 &    0.9780 \\
\rowcolor{Gray}
4  &         abalone &   0.1290 &  0.3745 &  0.3520 &  0.3286 &    0.2297 &    0.3830 \\
5  &  sick\_euthyroid &   0.8536 &  0.8337 &  0.8348 &  0.8463 &    0.8584 &    0.8070 \\
\rowcolor{Gray}
6  &    spectrometer &   0.8197 &  \textbf{0.8676 }&  0.8236 &  0.7643 &    0.7990 &    0.7462 \\
7  &     car\_eval\_34 &   0.8902 &  0.9041 &  0.9179 &  0.9278 &    0.9313 &    0.7537 \\
\rowcolor{Gray}
8  &        us\_crime &   0.4584 & \textbf{ 0.5284} &  0.5037 &  0.4853 &    0.4684 &    0.4516 \\
9  &       yeast\_ml8 &   0.0000 &  0.0284 &  0.0087 &  0.0386 &    0.0000 &    0.1606 \\
10 &           scene &   0.0095 &  0.1784 &  0.1490 &  0.1418 &    0.0735 &    0.2597 \\
\rowcolor{Gray}
11 &     libras\_move &   0.6370 &  \textbf{0.8407} &  0.7863 &  0.7198 &    0.7128 &    0.5357 \\
12 &    thyroid\_sick &   0.8462 &  \textbf{0.8706} &  0.8531 &  0.8621 &    0.8706 &    0.7434 \\
\rowcolor{Gray}
13 &       coil\_2000 &   0.0345 &  0.0775 &  0.0828 &  0.0765 &    0.1195 &    0.1913 \\
14 &      arrhythmia &   0.0000 &  \textbf{0.4758} &  0.3086 &  0.2705 &    0.2381 &    0.2674 \\
\rowcolor{Gray}
15 &  solar\_flare\_m0 &   0.0542 &  0.1038 &  0.0923 &  0.1812 &    0.1309 &    0.2575 \\
16 &             oil &   0.3717 &  0.5867 &  0.5222 &  0.6136 &    0.4779 &    0.2640 \\
17 &      car\_eval\_4 &   0.8728 &  0.9205 &  0.9492 &  0.9455 &    0.9325 &    0.6355 \\
\rowcolor{Gray}
18 &    wine\_quality &   0.1848 &  \textbf{0.3862} &  0.3101 &  0.2905 &    0.2319 &    0.2277 \\
19 &      letter\_img &   0.9493 &  0.9571 &  0.9520 &  0.9580 &    0.9575 &    0.7339 \\
\rowcolor{Gray}
20 &       yeast\_me2 &   0.2617 &  \textbf{0.3944} &  0.3579 &  0.3697 &    0.2585 &    0.2281 \\
21 &     ozone\_level &   0.0267 &  \textbf{0.3751} &  0.2806 &  0.2823 &    0.1470 &    0.1938 \\
\rowcolor{Gray}
22 &     mammography &   0.6709 &  0.6188 &  0.6541 &  0.6140 &    0.6881 &    0.3856 \\
23 &      abalone\_19 &   0.0000 &  \textbf{0.0809} &  0.0491 &  0.0478 &    0.0000 &    0.0408 \\
\bottomrule
\end{tabular}
\end{table}
The results of the AUC measurements are presented in Table \ref{auc}. The AUC represents the likelihood that a randomly chosen positive  sample will be As shown in the table, the proposed method often outperforms the benchmark techniques. Concretely, the proposed method produces the top AUC value in 10 out of 24 cases. For comparison the SMOTE algorithm produces the highest AUC value just once. The new sampling technique produces strong results across different class ratios and fields of application. The overall results show that the proposed sampling method outperforms the benchmark mtheods.  

\begin{table}[h!]
\centering
\caption{Classification performance - as measured by AUC - using various sampling techniques.}
\label{auc}
\begin{tabular}{llrrrrrr}
\toprule
{ID} &            Name &  Initial &   Gamma &   SMOTE &  ADASYN &  ROS &  RUS \\
\midrule
0  &           ecoli &   0.4864 &  0.4770 &  0.4446 &  0.4365 &    0.2891 &    0.4174 \\
1  &  optical\_digits &   0.8023 &  \textbf{0.8929} &  0.8689 &  0.8525 &    0.8618 &    0.8121 \\
2  &        satimage &   0.4570 &  0.4597 &  0.4884 &  0.4979 &    0.5152 &    0.3810 \\
3  &      pen\_digits &   0.9625 & \textbf{0.9737} &  0.9678 &  0.9620 &    0.9677 &    0.9579 \\
4  &         abalone &   0.1138 &  0.2002 &  0.1835 &  0.1705 &    0.1300 &    0.2203 \\
5  &  sick\_euthyroid &   0.7487 &  0.7089 &  0.7142 &  0.7320 &    0.7529 &    0.6695 \\
6  &    spectrometer &   0.7054 &  \textbf{0.7731} &  0.7034 &  0.6172 &    0.6769 &    0.6001 \\
7  &     car\_eval\_34 &   0.8093 &  0.8305 &  0.8524 &  0.8684 &    0.8753 &    0.6083 \\
8  &        us\_crime &   0.3036 &  \textbf{0.3190} &  0.3034 &  0.2816 &    0.2937 &    0.2785 \\
9  &       yeast\_ml8 &   0.0736 &  0.0758 &  0.0739 &  0.0766 &    0.0736 &    0.0858 \\
10 &           scene &   0.0742 &  0.1194 &  0.0900 &  0.0919 &    0.0863 &    0.1351 \\
11 &     libras\_move &   0.5399 & \textbf{ 0.7532} &  0.7128 &  0.6197 &    0.5839 &    0.3504 \\
12 &    thyroid\_sick &   0.7373 &  0.7660 &  0.7457 &  0.7559 &    0.7719 &    0.5870 \\
13 &       coil\_2000 &   0.0685 &  0.0705 &  0.0738 &  0.0711 &    0.0757 &    0.0957 \\
14 &      arrhythmia &   0.0553 &  \textbf{0.3479} &  0.2776 &  0.2299 &    0.2364 &    0.1449 \\
15 &  solar\_flare\_m0 &   0.1414 &  0.1473 &  0.1458 &  0.1741 &    0.1496 &    0.1595 \\
16 &             oil &   0.2446 &  0.3705 &  0.3757 &  0.4189 &    0.3559 &    0.1397 \\
17 &      car\_eval\_4 &   0.7777 &  0.8546 &  0.9075 &  0.9016 &    0.8749 &    0.4693 \\
18 &    wine\_quality &   0.1158 &  \textbf{0.1791} &  0.1288 &  0.1168 &    0.1104 &    0.1108 \\
19 &      letter\_img &   0.9063 &  0.9179 &  0.9121 &  0.9212 &    0.9205 &    0.5792 \\
20 &       yeast\_me2 &   0.1578 &  \textbf{0.2022} &  0.1794 &  0.1692 &    0.1383 &    0.1201 \\
21 &     ozone\_level &   0.0361 &  \textbf{0.1771} &  0.1350 &  0.1115 &    0.0681 &    0.0981 \\
22 &     mammography &   0.4942 &  0.4138 &  0.4440 &  0.3979 &    0.4955 &    0.2209 \\
23 &      abalone\_19 &   0.0093 &  \textbf{0.0193} &  0.0125 &  0.0136 &    0.0077 &    0.0187 \\
\bottomrule
\end{tabular}
\end{table}


\section{Conclusion}
Imbalanced class distribution is an issue that appears in a number of fields including text classification, medical diagnostics, and fraud detection. Prevalence of one class over another results in classification bias. Although there exists a plethora of approaches to address this issue there remains a lot of room for improvement.
In this paper we present a new oversampling technique based on the gamma distribution. The proposed approach has a number of advantages over the existing sampling techniques. First, the gamma distribution naturally leads to the new minority points being  generated near the existing minority points. Second, the new points are created both in  the same and opposite directions from the neighboring minority points. Finally, the performance of the new sampling method can be further improved by tuning the parameters of the gamma distribution. 

The efficacy of the proposed method was tested on a range of simulated and real life datasets. We constructed a synthetic dataset to illustrate a scenario where the new sampling method produces optimal results. In general, the new method performs well in settings where the regions between neighboring minority points contain majority points. The use of gamma distribution helps to avoid generating the new minority points in mix with the majority points. The experimental results on the synthetic dataset show that the proposed sampling approach performs well regardless of the classification algorithm (Table \ref{syn_table}).

To validate the new sampling approach we carried out extensive tests on a wide array of real life datasets. We employed a total of 24 datasets from various fields of application and a range of class ratios. The proposed sampling technique produced the top $F_1$-scores on 12 (half) of the datasets (Table \ref{f1}). For comparison, the SMOTE algorithm produced the top score only on 1 dataset. Similar results were obtained when employing AUC to measure the classification performance. The results on the real life datasets indicate that the new sampling technique significantly outperforms other benchmark methods.

In this paper, we introduced a new sampling technique to deal with imbalanced data. The new method is based on the gamma distribution and allows for generation of minority points in the vicinity of existing points. The sampling procedure can be fine tuned by adjusting the distribution parameters. The results of extensive experiments show that the proposed method significantly outperforms other benchmark methods. Therefore, we believe that it could be a valuable tool for researchers working with imbalanced data.


\end{document}